\documentclass[11pt,a4paper]{article}
\usepackage[hyperref]{acl2020}

\aclfinalcopy 

\usepackage{times}
\usepackage{latexsym}
\usepackage{url}

\usepackage[algoruled,linesnumbered,ruled,vlined]{algorithm2e}
\usepackage{times}
\usepackage{latexsym}
\usepackage{pgfplots}
\usepackage{graphicx}
\usepackage{enumitem}
\usepackage{xcolor,colortbl}
\usepackage{tikz}
\usepackage{diagbox}
\usepackage{tkz-tab}
\usepackage{caption}
\usepackage{latexsym}
\usepackage{amssymb}
\usepackage{amsmath}
\usepackage{subcaption}
\usepackage{natbib}
\usepackage{blindtext}
\usepackage{url}
\usepackage{color}

\usepackage{url}
\usepackage{hyperref}    
\usepackage{amsthm}   
\usepackage{cleveref}
\SetAlFnt{\small}
\SetAlCapFnt{\small}
\SetAlCapNameFnt{\small}

\def\aclpaperid{44} 

\usepackage[para,multiple]{footmisc}

\title{On Dimensional Linguistic Properties of the Word Embedding Space}

\author{
  Vikas Raunak\thanks{equal contribution} \\
  Carnegie Mellon University \\
  {\tt vraunak@cs.cmu.edu } \And
     Vaibhav Kumar$^{*}$ \\
  Carnegie Mellon University \\
  {\tt vaibhav2@cs.cmu.edu } \AND
  Vivek Gupta \\
  University of Utah  \\
  {\tt vgupta@cs.utah.edu} \And
  Florian Metze\\
  Carnegie Mellon University \\
  {\tt fmetze@cs.cmu.edu} \\
  }

\date{}

\begin{document}

\maketitle
\begin{abstract}
Word embeddings have become a staple of several natural language processing tasks, yet much remains to be understood about their properties. In this work, we analyze word embeddings in terms of their principal components and arrive at a number of novel and counterintuitive observations. In particular, we characterize the utility of variance explained by the principal components as a proxy for downstream performance. Furthermore, through syntactic probing of the principal embedding space, we show that the syntactic information captured by a principal component does not correlate with the amount of variance it explains. Consequently, we investigate the limitations of variance based embedding post-processing, used in a few algorithms such as \cite{mu2018allbutthetop, raunak} and demonstrate that such post-processing is counter-productive in sentence classification and machine translation tasks. Finally, we offer a few precautionary guidelines on applying variance based embedding post-processing and explain why non-isotropic geometry might be integral to word embedding performance.
\end{abstract}

\section{Introduction}
\label{sec:1}
Word embeddings have revolutionized natural language processing by representing words as dense real-valued vectors in a low dimensional space. Pre-trained word embeddings such as Glove \cite{pennington2014glove}, word2vec \cite{mikolov2013distributed} and  fastText \cite{bojanowski2017enriching}, trained on large corpora are readily available for use in a variety of tasks. Subsequently, there has been emphasis on post-processing the embeddings to improve their performance on downstream tasks \cite{mu2018allbutthetop} or to induce linguistic properties \cite{mrkvsic2016counter, faruqui2015retrofitting}. In particular, the Principal Component Analysis (PCA) based post-processing algorithm proposed by \cite{mu2018allbutthetop} has led to significant gains in word and sentence similarity tasks, and has also proved useful in dimensionality reduction \cite{raunak}. Similarly, understanding the geometry of word embeddings is another area of active research \cite{mimno2017strange}. In contrast to previous work such as \cite{dim}, which focuses on  optimal dimensionality selection for word embeddings, we explore the dimensional properties of existing pre-trained word embeddings through their principal components. Specifically, our contributions are as follows:
\begin{enumerate}
\setlength\itemsep{0.25em}
\item We analyze the word embeddings in terms of their principal components and demonstrate that their performance on both word similarity and sentence classification tasks saturates well before the full dimensionality.
\item We demonstrate that the amount of variance captured by the principal components is a poor representative for the downstream performance of the embeddings constructed using the very same principal components.
\item We investigate the reasons behind the aforementioned result through syntactic information based dimensional linguistic probing tasks \cite{conneau2018you} and demonstrate that the syntactic information captured by a principal component is independent of the amount of variance it explains.
\item We point out the limitations of variance based post-processing used in a few algorithms \cite{mu2018allbutthetop, raunak} and demonstrate that it leads to a decrease in performance in sentence classification and machine translation tasks, restricting its efficacy mainly to semantic similarity tasks.
\end{enumerate}

In Section \ref{sec:1}, we provide an introduction to the problem statement. In Section \ref{sec:2}, we discuss the dimensional properties of word embeddings. In Section \ref{sec:3}, we conduct a variance based analysis by evaluating the word embeddings on several downstream tasks. In Section \ref{sec:4}, we move on to dimensional linguistic probing tasks followed by Section \ref{sec:5}, where we discuss variance based post-processing algorithms, and finally conclude in Section \ref{sec:6}. To foster reproducibility, we have released the source code along with paper \footnote{\url{https://github.com/vyraun/dlp}}.

\begin{figure}
    \includegraphics[height=5.0cm, width=0.48\textwidth]{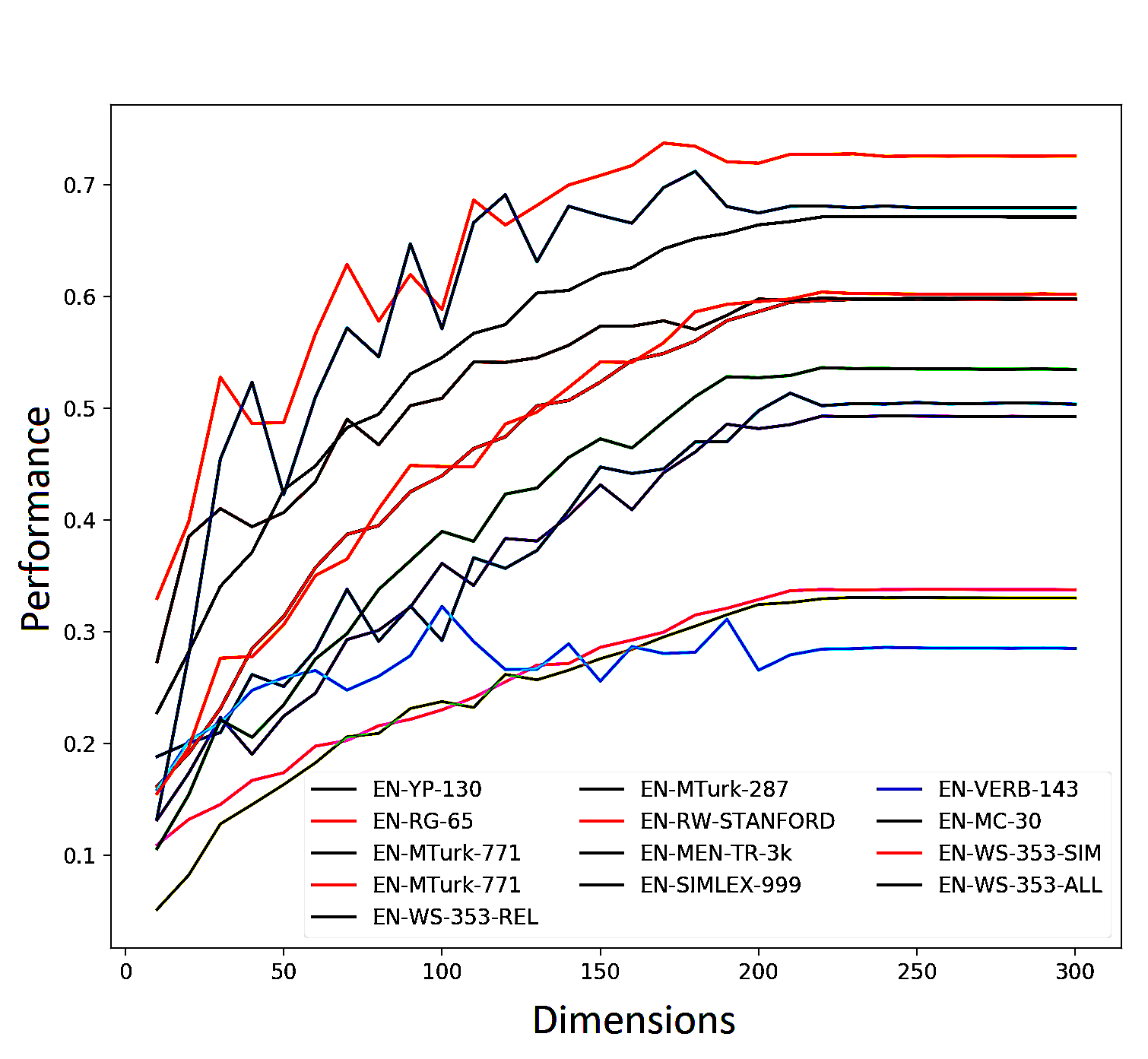}
    \caption{\textbf{Rho} x 100 on Word Similarity Tasks}
    \label{fig:var_tasks}
\end{figure}
\begin{figure}
    \includegraphics[height=5.0cm,width=0.48\textwidth]{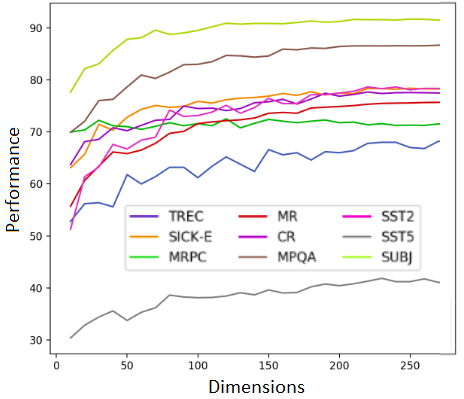}
    \caption{Accuracy on Sentence Classification Tasks}
    \label{fig:sem_eval_task_acc}
    \vspace{-1.0em}
\end{figure}

\section{Dimensional Properties of the Word Embedding Space}
\label{sec:2}
Principal components provide a natural basis for studying the properties of an embedding space. In this work, we refer to the properties pertaining to the principal components of the embedding space as dimensional properties and the embedding space obtained by projecting the embeddings on the principal components as the principal embedding space. We study the principal embedding space and the dimensional properties in a number of different contexts such as word similarity, sentence classification. We provide a brief introductions to the both evaluation tasks of our experiment in the sub-sections. For details on the benchmarks, please refer to \citet{conneau2018senteval} for sentence classification and \citet{faruqui2014community} for word similarity.

\vspace{0.5em}
For experiments in this section, we use $300$ dimensional a) Glove embeddings (trained on Wikipedia $201$ + Gigaword $5$ \footnote{\url{https://stanford.io/2Gdv8uo}}), b) fastText embeddings (trained on Wikipedia, UBMC webbase corpus and statmt.org news dataset \footnote{\url{https://bit.ly/2FMTB4N}}) and c) Word2vec  embeddings (trained on the GoogleNews dataset \footnote{\url{https://bit.ly/2esteWf}}. We use Glove embeddings for the word similarity tasks. For the sentence classification tasks, we show results for fasttext and word2vec as well, in addition to Glove embeddings. For the sentence classification tasks we use Logistic Regression as the classifier, since it is the simplest classification model and we are only interested in evaluating performance variation due the changes in representations. Thus, the convex objective used in the classifier avoids any optimizer instability, making our entire evaluation pipeline deterministic and exactly reproducible.

\begin{table*}
\centering
\caption{Test accuracy of embeddings composed of \textbf{Top-100 (T)}, \textbf{Middle-100 (M)} and \textbf{Bottom-100 (B)} \textbf{principal components} on sentence classification datasets. The highlighted cells correspond to one of the three cases - \emph{\textbf{M outperforms T}} (\colorbox{orange!50}{\textbf{orange}}),  \emph{\textbf{B outperforms T}} (\colorbox{red!50}{\textbf{red}}) and  \emph{\textbf{B outperforms M}} (\colorbox{yellow}{\textbf{yellow}})}.
\scalebox{0.95}{
\begin{tabular}{c|c|c|c|c|c|c|c|c|c}
\hline
Split & \textbf{MR}    & \textbf{CR}    & \textbf{SUBJ}  & \textbf{MPQA}  & \textbf{SST2}  & \textbf{SST5}  & \textbf{TREC} & \textbf{SICK-E} & \textbf{MRPC}  \\ \hline
\textbf{Random-Embeddings}       & 61.65 & 71.6 & 78.9  & 73.79 & 60.57 & 31.09 & 70.0 & 77.07  & 69.91 \\ \hline
\textbf{Glove-Full}       & 75.7 & 77.48 & 91.76  & 86.66 & 78.03 & 41.0 & 68.0 & 78.49  & 70.61 \\ 
\textbf{Glove-T}       & 70.74 & 73.67 & 90.1  & 81.58 & 72.49 & 37.24 & 61.8 & 75.71  & 71.94 \\ 
\textbf{Glove-M}   & \cellcolor{orange!50}\textbf{72.98} & \cellcolor{orange!50}\textbf{75.04} & 87.76 & \cellcolor{orange!50}\textbf{84.07} & \cellcolor{orange!50}\textbf{75.34} & \cellcolor{orange!50}\textbf{40.5}  & 57.6 & \cellcolor{orange!50}\textbf{76.5}   & 71.42 \\ 
\textbf{Glove-B}  & 67.62 & 73.01 & 83.68 & \cellcolor{red!50}\textbf{81.61} & 69.52 & 36.11 & 57.0 & 72.82  & 70.96 \\ \hline
\textbf{Word2vec-Full}       & 77.65 & 79.26 & 90.76  & 88.3 & 79.68 & 42.44 & 83.0 & 78.24  & 72.58 \\ 
\textbf{Word2vec-T}  & 74.34 & 76.29 & 89.88 & 85.07 & 77.16 & 40.36 & 70.0 & 75.46  & 71.48 \\ 
\textbf{Word2vec-M} & 72.91 & 73.43 & 82.39 & 82.76 & 72.65 & 38.69 & 66.0 & 70.53  & 71.36 \\ 
\textbf{Word2vec-B} & 71.42 & \cellcolor{yellow!50}\textbf{74.25} & \cellcolor{yellow!50}\textbf{82.47} & 81.05 & \cellcolor{yellow!50}\textbf{73.48} & 38.46 & \cellcolor{red!65}\textbf{72.2} & \cellcolor{yellow!50}\textbf{74.3}   & 71.01 \\ \hline
\textbf{fastText-Full}       & 67.85 & 75.39 & 85.87 & 79.85 & 70.57 & 35.97 & 68.0 & 76.66  & 70.84 \\ 
\textbf{fastText-T}  & 69.42 & 67.76 & 87.69 & 84.64 & 74.35 & 36.83 & 74.8 & 66.04  & 70.61 \\ 
\textbf{fastText-M} & 68.88 & 65.3  & 81.74 & 81.45 & 72.1  & 35.57 & 65.2 & 65.01  & 68.29 \\ 
\textbf{fastText-B} & 66.45 & 64.21 & 79.89 & 79.83 & 69.96 & 31.22 & \cellcolor{yellow!50}\textbf{69.4} & 63.77  & 67.94 \\ \hline
\end{tabular}}
\label{tab:sem_eval}
\vspace{-1.00em}
\end{table*}

\subsection{Word Similarity Tasks}
\label{sec:2.1}
The word similarity benchmarks \cite{faruqui2014community} have word pairs (WP) that have been assigned similarity rating by humans.  While evaluating word embeddings, the similarity between the words is calculated by the cosine similarity of their vector representations. Then, Spearman’s rank correlation co-efficient (\textit{Rho}) between the ranks produced using the cosine similarities and the given human rankings is used for the performance evaluation. Hence, for better word similarity,  the evaluation metric (\textit{Rho}) will be higher.

\vspace{0.5em}
Figure \ref{fig:var_tasks} shows the performance (\textit{Rho} x 100) of word embeddings (Glove) on $13$ word similarity benchmarks w.r.t varying word embedding dimensions. The similarities are computed by projecting the embeddings in the principal component space. Each new evaluation cumulatively adds $10$ more principal components to the earlier embeddings, i.e. the units on the X-axis vary in the increments of 10. Thus, we obtain $30$ measurements for each dataset, ranging from word embeddings constructed using the first $10$ principal components to orignal $300$ principal components. From Figure \ref{fig:var_tasks}, it is evident that the performance saturates consistently at around $200$ dimensions for all of the tasks, after which adding new principal components does not lead to much gain in performance.

\subsection{Sentence Classification Tasks}
\label{sec:2.2}
The sentence classification tasks \cite{conneau2018senteval} include binary classification tasks (MR, CR, SUBJ, MPQA), multiclass classification tasks (SST-FG,  TREC),  entailment  (SICK-E),  semantic relatedness (STS-B) and Paraphrase detection (MRPC) tasks. As usual, the evaluation is done by computing the classification accuracy on the test set.

\begin{table*}
\centering
\caption{Performance on sentence classification tasks of various embeddings (300 dimensional) and their post-processed PCA reduced counterparts of 200 dimensions.}
\scalebox{0.95}{
\begin{tabular}{c|c|c|c|c|c|c|c|c|c}
\hline
Embedding & \textbf{MR}    & \textbf{CR}    & \textbf{SUBJ}  & \textbf{MPQA}  & \textbf{SST2}  & \textbf{SST5}  & \textbf{TREC} & \textbf{SICK-E} & \textbf{MRPC}  \\ \hline
\textbf{Glove}    & 75.7  & 77.48 & 91.76 & 86.66 & 78.03 & 41.0  & 68.0 & 78.49  & 70.61 \\ 
\textbf{Glove-PCA}       & 74.62 & 76.95 & 91.6 & 85.97 & 77.16 & 40.18 & 66.6 & 77.02  & 72.99 \\ 
\textbf{Glove-200}       & 74.69 & 77.91 & 91.18 & 86.52 & 77.98 & 40.05 & 66.4 & 77.47  & 72.23 \\ \hline
\textbf{Word2vec} & 77.65 & 79.26 & 90.76 & 88.30 & 79.68 & 42.44 & 83.0 & 78.24  & 72.58 \\
\textbf{Word2vec-PCA}    & 76.53 & 78.12  & 90.50 & 86.74 & 79.63 & 41.49 & 77.6 & 76.54  & 72.17 \\ \hline
\textbf{fastText} & 67.85 & 75.39 & 85.87 & 79.85 & 70.57 & 35.9 & 68.0 & 76.66  & 70.84 \\
\textbf{fastText-PCA}    & 66.83 & 74.46 & 85.26  & 78.91  & 69.85 & 36.11 & 66.0 & 76.50  & 68.75 \\ \hline
\end{tabular}}
\label{tab:pca_200}
\vspace{-1.00em}
\end{table*}

\vspace{0.5em}
Figure \ref{fig:sem_eval_task_acc} shows the performance (Test accuracy) on $9$ standard downstream sentence classification tasks \cite{conneau2018senteval} using the same procedure for constructing word embeddings (Glove) as in \ref{sec:2.1}. Further, sentence vectors were constructed using an average of the contained word embeddings, which has been demonstrated to be a very strong baseline for downstream tasks \cite{arora2016simple}. From Figure \ref{fig:sem_eval_task_acc}, we can observe that, similar to the previous word similarity tasks, the performance saturates consistently at around $200$ dimensions for all of the tasks, after which incrementing the embeddings with additional principal components does not lead to much gains in performance. We also report results for original $(300D$) and post processed PCA reduced ($200D$) word embeddings for other types (fastText, Glove) in Table \ref{tab:pca_200}. In Table \ref{tab:pca_200}, we also report results with pretrained $200D$ Glove embedding. \footnote{word embeddings for $200D$ for other embedding types (fasttext, word2vec) are not publicly available.} 

\vspace{0.5em}
\noindent \textbf{Analysis: }To conclude, observations from both word similarity and sentence classification tasks, of saturation in performance around $200$, much before the original $300$ dimensions implies redundancy among the dimensions (in section \ref{sec:3} we will clarify why it doesn't imply \textit{noise}). Furthermore, this observation is consistent across various embedding types (Glove, fastText and word2vec) for the sentence classification tasks, as demonstrated in Table \ref{tab:pca_200}. This also suggests a simple strategy to reduce the embedding size wherein one third of the components could be reliably removed without affecting the performance on word similarity or sentence classification tasks, leading to $~33\%$ memory reduction.

\section{Variance Based Analysis}
\label{sec:3}
In this section, we characterize the redundancy observed in Section \ref{sec:2}, in terms of variance of the principal components. Specifically, we measure downstream performance (on the sentence classification tasks of Section \ref{sec:2.2}) of word embeddings against the amount of variance captured by the principal components (the variance explained or captured by a principal component is the variance of the embeddings when projected onto that principal component; hereon, we refer to the fraction of variance explained by a principal component simply as variance explained by that component). Similar to the previous section, we use $300$ dimensional Glove embeddings (trained on Wikipedia $201$ + Gigaword $5^2$) for experiments in this section, along with publically released fastText (trained on Wikipedia, UBMC webbase corpus and statmt.org news dataset$^3$ and Word2vec (trained on the GoogleNews dataset$^4$) embeddings, both of 300 dimensions.

\begin{table}
\centering
\caption{\small The Variance for each of the T, M, B splits of the embeddings.}
\scalebox{0.9}{
\begin{tabular}{c|c|c|c}
\hline
& \textbf{Glove} & \textbf{Word2vec} & \textbf{fastText} \\ \hline
\textbf{T}   & 0.529 & 0.628   & 0.745      \\ \hline
\textbf{M}  & 0.371 & 0.221   & 0.162 \\ \hline
\textbf{B}   & 0.100  & 0.151  & 0.093 \\ \hline
\end{tabular}}
\label{tab:variance_splits}
\vspace{-1.25em}
\end{table}

\vspace{0.5em}
For each of the embedding types, we first construct word embeddings using only top $100$ principal components (T), the middle $100$ principal components (M) and the bottom $100$ principal components (B). Then, we compute the variance for each split by aggregating the variance of the $100$ principal components of each split for all three embedding types. Table \ref{tab:variance_splits} highlights how the total variance is divided across the three splits. The T embeddings have the first $100$ principal components (PCs), so the highest variance explained, while the $B$ embeddings have the bottom 100 components, thereby the least variance explained. Furthermore, the variance explained by the principal components for the same split also differ significantly across the different embedding types. For example, fastText has more variance explained, when compared to Glove and Word2vec, for the split T, while Glove has the most variance explained, among the three embedding types, for the split M. Lastly, Word2vec explains more variance than Glove and fastText for the split B. The differences are expected since, the three embedding types differ considerably in their training algorithms. While Word2vec uses negative sampling, Glove derives semantic relationships from the word-word co-occurrence matrix and fastText uses subword information. So, to summarize we constructed altogether $9$ embedding splits ($3$ from each of the $3$ embedding types), which differ significantly in terms of the variance explained by their constituent components.

We use the $100$ dimensional embedding obtain from the several splits (T, M, B) and types (Glove, fastText, Word2vec) as features for downstream sentence classification tasks, as in Section \ref{sec:2.2}, except that, now, each of the embedding feature has $100$ dimensions. The experiments are designed to test whether the variance explained by a split is closely correlated with the downstream performance metric (classification accuracy) for each of the three embedding types. Table \ref{tab:sem_eval} shows the results on $9$ sentence classification tasks, for each embedding split, for all the three embedding types. In the table, the highlighted cells represent the cases where classification accuracy of the lower variance split exceeds that of the corresponding higher variance split. Each annotated cell corresponds to one of the three cases - \textbf{M outperforms T} (\colorbox{orange!50}{\textbf{orange}}), \textbf{B outperforms T} (\colorbox{red!50}{\textbf{red}}) and  \textbf{B outperforms M} (\colorbox{yellow}{\textbf{yellow}}). For the comparisons between T and M splits, in 6 out of 27 such comparisons, the M embeddings outperform the T embeddings. Similarly, for comparisons between M and B embeddings, the B embeddings outperform the M embeddings in $7$ out of $27$ cases and for comparisons between T and B embeddings, in $2$ out of $27$ cases the B embeddings outperform the T embeddings. Further, in a number of cases (although not highlighted), such as on the MRPC task, the T and M splits differ very little in performance. The same is true for M and B splits on tasks such as MPQA and CR. Such cases are least prominent in fastText, probably due to the extremely large gap in the variance explained between the T, M and T, B splits.\\

\noindent \textbf{Analysis: } From Table \ref{tab:sem_eval} it is evident that the performance drop between the T, M, B splits is quite low for a number of tasks, which is highly contrary to the expectation, given the large differences in the variance explained (Table \ref{tab:variance_splits}). Further, there are also many cases where lower variance embeddings (B and M) outperform the emedddings (M and T) with higher variance, for all the three embedding types. These results demonstrate that for word embeddings, \emph{the variance explained by the principal components is not sufficient for explaining their downstream performance. In other words, the variance explained by the principal components is a weak representative of downstream performance.} This is in contrast to the widely used practice of using the variance explained by the principal components as a fundamental tool to assess the quality of the corresponding representations \cite{jolliffe2016principal}. 

\begin{figure}
\centering
\begin{subfigure}[b]{0.45\textwidth}
\centering
\includegraphics[height=5.0cm, width=\textwidth]{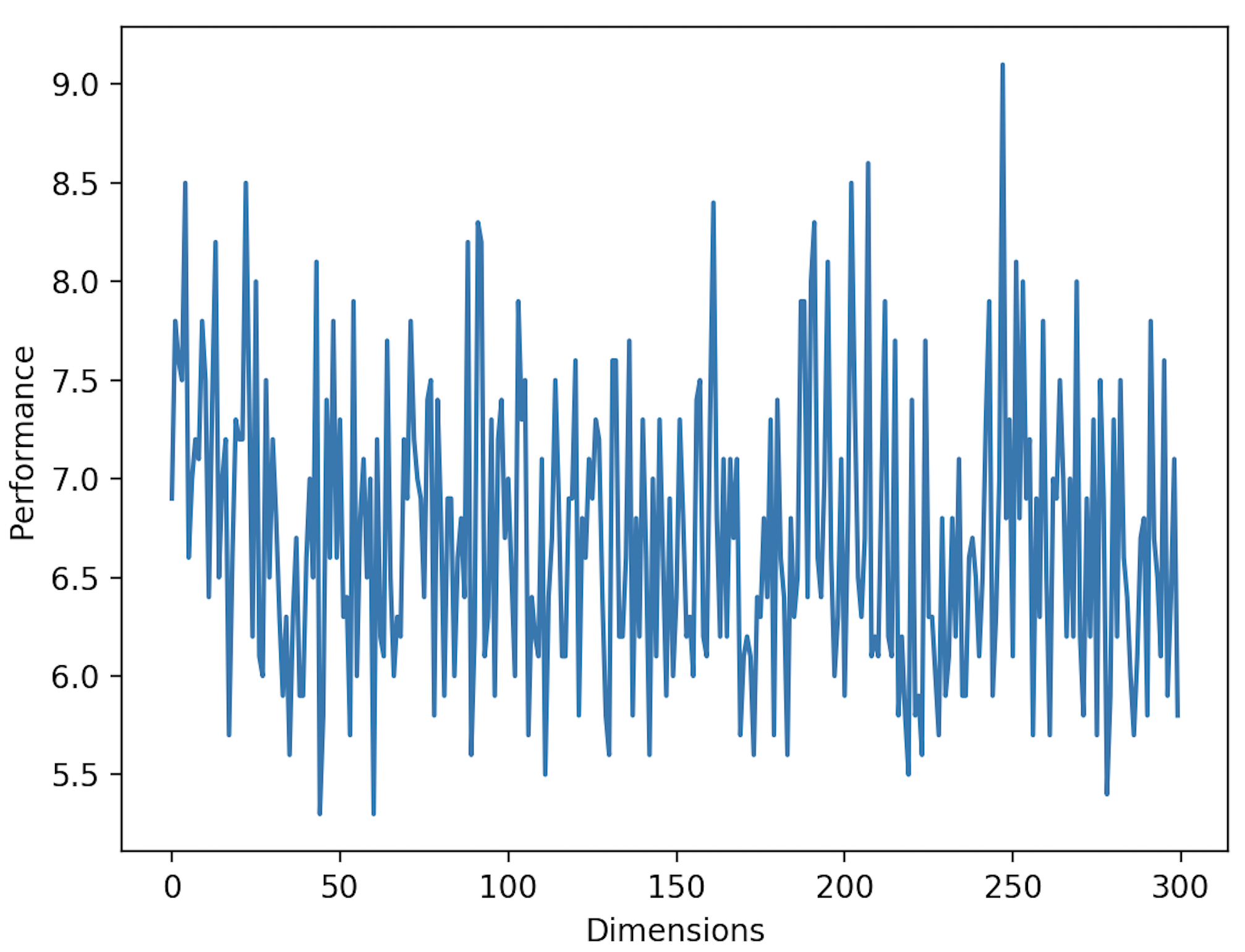}
\end{subfigure}
\begin{subfigure}[b]{0.45\textwidth}
\centering
\captionsetup{font=small, skip=0pt}
\includegraphics[height=5.0cm, width=\textwidth]{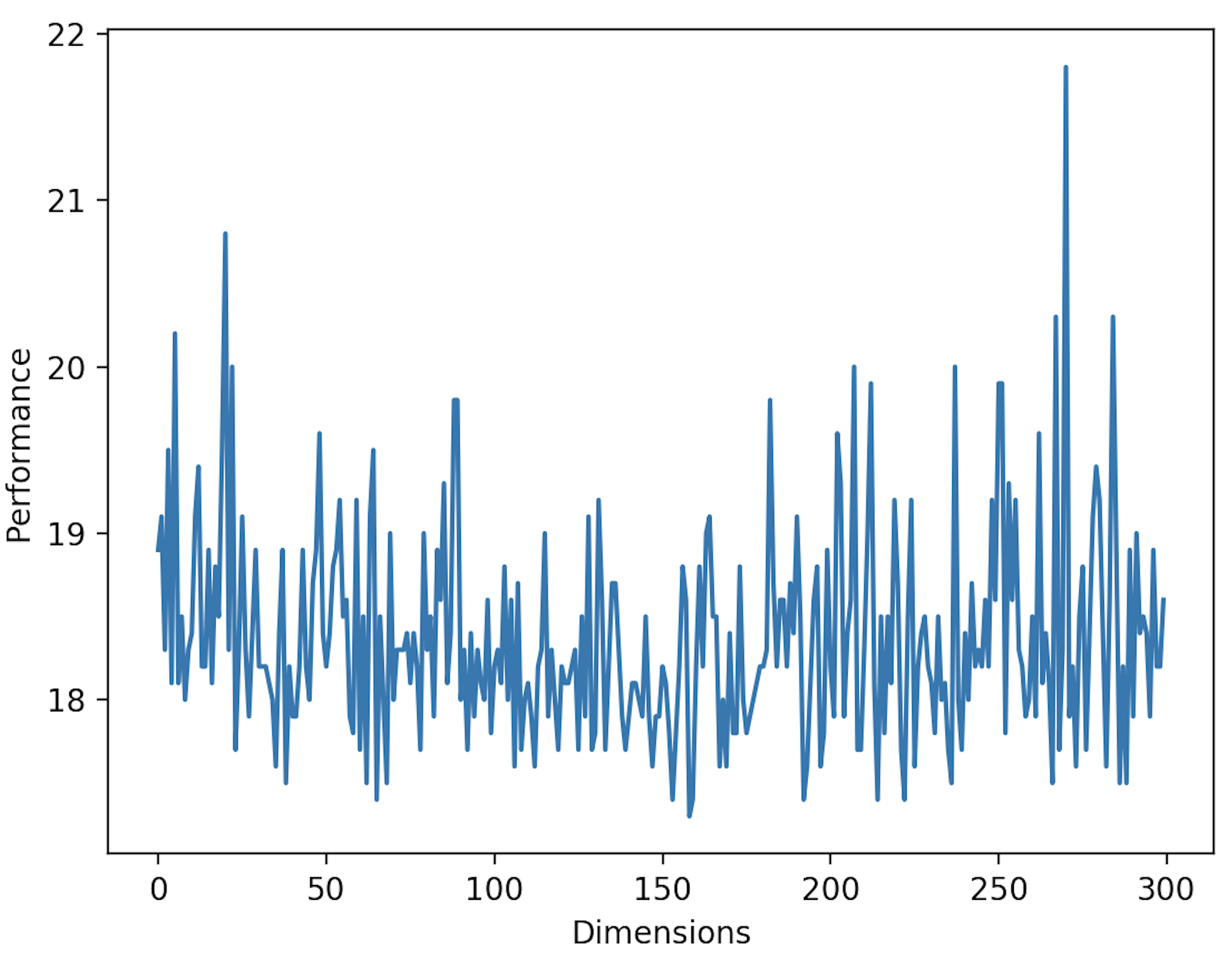}
\end{subfigure}
\caption{Analysis of individual principal components on the two syntactic information based linguistic probing tasks: TopConst (top) and TreeDepth (bottom). The Y-axis represents the Test accuracy on the two tasks.}
\label{fig:full_an}
\vspace{-1.25em}
\end{figure}

\section{Dimensional Linguistic Probing Tasks}
\label{sec:4}
A plausible hypothesis to explain the better performance of M and B embeddings (Table \ref{tab:sem_eval}) in the earlier section is that \emph{`the syntactic information required for downstream sentence classification tasks is distributed independently with respect to the principal components'}. To explore the validity of the proposed hypothesis, we leverage two linguistic probing tasks, namely TreeDepth and TopConst \cite{conneau2018you}. These probing tasks are designed to test whether sentence embeddings are sensitive to the syntactic properties of the encoded sentences. The TreeDepth task (a 8-way classification problem) tests whether the model can predict the depth of the hierarchical syntactic structure of the sentence. For doing well on the TreeDepth task, the embeddings have to group sentences by the depth of the longest path from root to any leaf. In the TopConst task (a 20-way classification problem), a sentence must be classified in terms of the sequence of its constituents occurring immediately below the sentence node of its hierarchical structure. Therefore, for good performance on the TopConst task, the embeddings have to capture latent syntactic  structures and cluster  them  by  constituent  types. The random baselines for the TreeDepth and TopConst tasks are $12.5$ and $5.0$ respectively, while full 300-dimensional Glove embeddings obtain accuracies of $37$ and $68$ percent respectively. 

\vspace{0.5em}
To evaluate the syntactic information contained in each of the principal components, we first construct one-dimensional word embeddings by projecting word vectors onto a single principal component. Then we use these word embeddings to construct sentence vectors, as in Section \ref{sec:2.2}, which are used as features for the two classification tasks. For good performance, the single component sentence vector has to distinguish between the probing task’s output classes. Therefore, the performance on these tasks can be used to isolate the behavior of individual components with respect to the syntactic information captured. The motivation here is that if the syntactically discriminative components would vary considerably, then we can isolate the behavior of the individual components and see their correspondence with the rank of the principal component. Figure \ref{fig:full_an} depicts the scores (Test classification accuracy) on TopConst and TreeDepth tasks respectively.  The average performance of the one-dimensional representations has mean $\pm$ standard deviation of $18.38 \pm 0.64$ and $6.71 \pm 0.72$ for the TreeDepth and TopConst tasks respectively. 

\vspace{0.5em}
\noindent \textbf{Analysis: }The average performance of the one-dimensional representations on both tasks is much lower than full dimension embeddings but well above the random baseline. However, many individual compoenents far exceed the random baseline as well. As mentioned earlier, we wanted to probe whether such discriminativeness is ranked according to variance. However from Figure \ref{fig:full_an}, it is evident that the performance across the dimensions does not have any particular trend (increasing or decreasing) w.r.t to the rank of the principal components. In fact, the peak performance on both the tasks is achieved by a component in the bottom (B) split of the embeddings. This validates the hypothesis that \emph{the syntactic information captured by a principal component is independent of the amount of variance it explains}.

\begin{table}[!htbp]
\centering
\caption{Classification Accuracy for Linguistic Probing Tasks using the \textbf{T}, \textbf{M}, \textbf{B} splits of the embeddings. Here also, the highlighted cells correspond to one of the three cases - \emph{\textbf{M outperforms T}} ( \colorbox{orange!50}{\textbf{orange}}),  \emph{\textbf{B outperforms T}} (\colorbox{red!50}{\textbf{red}}) and  \emph{\textbf{B outperforms M}} (\colorbox{yellow}{\textbf{yellow}})}
\scalebox{0.9}{
\begin{tabular}{c|c|c}
\hline
Embedding & \textbf{TopConst} & \textbf{TreeDepth} \\ \hline
\textbf{Glove-T}   & 28.1 & 28.2  \\ 
\textbf{Glove-M}   & 26.0 & 24.8  \\ 
\textbf{Glove-B}   & \cellcolor{yellow!50}\textbf{27.1} & \cellcolor{yellow!50}\textbf{26.9}  \\ \hline
\textbf{Word2vec-T}  & 23.9 & 42.5 \\ 
\textbf{Word2vec-M}  & \cellcolor{orange!50}\textbf{24.3} & \cellcolor{orange!50}\textbf{43.5} \\ 
\textbf{Word2vec-B}  & 23.7 & \cellcolor{red!65}\textbf{44.6} \\ \hline
\textbf{fastText-T}   & 31.2  & 50.7 \\ 
\textbf{fastText-M}   & 29.3  & \cellcolor{orange!50}\textbf{51.0} \\ 
\textbf{fastText-B}   & \cellcolor{yellow!50}\textbf{30.6}  & \cellcolor{red!65}\textbf{56.8} \\ \hline
\end{tabular}}
\label{tab:probing}
\vspace{-0.5em}
\end{table}

To further validate the hypothesis, we repeat the experiment described in Section \ref{sec:3} for each of the embedding types, except on the synctatic probing tasks of TopConst and TreeDepth in Table \ref{tab:probing}. Similar to Table \ref{tab:sem_eval}, each annotated cell in Table \ref{tab:probing} corresponds to one of the three cases - \textbf{M outperforms T} (\colorbox{orange!50}{\textbf{orange}}), \textbf{B outperforms T} (\colorbox{red!50}{\textbf{red}}) and  \textbf{B outperforms M} (\colorbox{yellow}{\textbf{yellow}}). For the comparisons between T and M splits, in $3$ out of $6$ such comparisons, the M embeddings outperform the T embeddings. Similarly, for comparisons between M and B embeddings, the B embeddings outperform the M embeddings in $5$ out of $6$ cases and for comparisons between T and B embeddings, in $2$ out of $6$ cases the B embeddings outperform the T embeddings. In other words, table \ref{tab:probing} shows that for the TreeDepth task, the B embeddings significantly outperform T and M embeddings for word2vec and fastText, whereas for Glove, it outperforms the M embeddings. For the TopConst task as well, the B embeddings outperform M embeddings for Glove and fastText, whereas for Word2vec, it outperforms the T embeddings. Thus, the discrepancy in performance on these syntactic probing tasks is even more severe when compared to the sentence classification tasks evaluated in Section \ref{sec:3}. The results also validate our hypothesis that \emph{the variance explained by the embeddings is of little predictive strength in predicting its relative performance}.

\begin{table*}
\centering
\caption{Performance on sentence classification tasks of various embeddings and their post-processed (PPA) counterparts. The \colorbox{red!50}{\textbf{red}} colored cells denote the cases where the original embeddings outperformed their post-processed (PPA) counterparts.}
\scalebox{0.95}{
\begin{tabular}{c|c|c|c|c|c|c|c|c|c}
\hline
Embedding & \textbf{MR}    & \textbf{CR}    & \textbf{SUBJ}  & \textbf{MPQA}  & \textbf{SST2}  & \textbf{SST5}  & \textbf{TREC} & \textbf{SICK-E} & \textbf{MRPC}  \\ \hline
\textbf{Glove (300 dim)}    & 75.7  & 77.48 & 91.76 & 86.66 & 78.03 & 41.0  & 68.8 & 78.49  & 70.61 \\ 
\textbf{PPA on Glove}       & \cellcolor{red!50}\textbf{75.57} & 77.48 & \cellcolor{red!50}\textbf{91.01} & 86.67 & \cellcolor{red!50}\textbf{77.98} & \cellcolor{red!50}\textbf{40.72} & \cellcolor{red!50}\textbf{65.8} & 78.53  & 71.59 \\ \hline
\textbf{Word2vec (300 dim)} & 77.65 & 79.23 & 90.76 & 88.30 & 79.68 & 42.44 & 82.6 & 78.24  & 72.64 \\
\textbf{PPA on Word2vec}    & \cellcolor{red!50}\textbf{77.33} & 79.5  & \cellcolor{red!50}\textbf{90.59} & \cellcolor{red!50}\textbf{88.12} & \cellcolor{red!50}\textbf{79.41} & 42.71 & 83.4 & 78.26  & \cellcolor{red!50}\textbf{72.58} \\ \hline
\textbf{fastText (300 dim)} & 74.16 & 71.63 & 89.56 & 87.12 & 79.24 & 39.14 & 79.4 & 72.34  & 70.14 \\
\textbf{PPA on fastText}    & 74.59 & 71.63 & \cellcolor{red!50}\textbf{89.4}  & \cellcolor{red!50}\textbf{86.9}  & \cellcolor{red!50}\textbf{79.13} & 39.64 & 80.2 & 72.36  & \cellcolor{red!50}\textbf{70.09} \\ \hline
\end{tabular}}
\label{tab:ppa_vs_glo}
\vspace{-1.15em}
\end{table*}

\vspace{-0.5em}
\section{The Post Processing Algorithm (PPA)}
\label{sec:5}
\vspace{-0.25em}
In this section, we briefly describe and then evaluate the post-processing algorithm (PPA) by \cite{mu2018allbutthetop}, which achieves high scores on Word and Semantic textual similarity tasks \cite{agirre2012semeval}. The algorithm (PPA) is listed below as Algorithm \ref{algo:PPA}. PPA removes the projections of top principal components from each of the word vectors, making the individual word vectors more discriminative. The algorithm could be regarded as pushing the word embeddings towards a more isotropic space \cite{arora2016latent}, by eliminating the common parts (mean vector and top principal components of the embedding space) from the individual word embeddings. However, it is worth revisiting the assumption whether isotropy (or angular isotropy more specifically) of the embedding space is universally beneficial with respect to downstream tasks. In this section, we stress test this assumption on a range of sentence classification and machine translation tasks. Our fundamental intuition is that since these tasks require the embedding space to capture syntactic properties much more significantly than word-similarity tasks, enforcing isotropy could lead to worse performance.

\label{ssec:layout}
 \begin{algorithm}[h!]
     \SetAlgoNoLine
     \SetNoFillComment
     \KwData{Embedding Matrix X, Threshold Parameter D}
     \KwResult{Post-Processed Word Embedding Matrix X}
     \tcc{Subtract Mean Embedding}
     X = X - $\overline{\text{X}}$ \;
     \tcc{Compute PCA Components}
     $u_i$ = PCA(X), where i = $1$, $2$, $\ldots$ $d$ \;
     \tcc{Remove Top-D Components}
     \For{\textbf{all} v in X}{
     $ v = v -  \sum_{i=1}^{D}(u_i^T \cdot v)u_i $
     }
 \caption{Post Processing Algorithm PPA(X, D)}
 \label{algo:PPA}
 \end{algorithm}

\vspace{-0.5em}
\subsection{Sentence Classification Tasks}
\vspace{-0.25em}
We compare the performance of PPA (with a constant D=$5$ across all the embeddings) on the $9$ downstream sentence classification tasks, as in Section \ref{sec:3}. The results are presented in Table \ref{tab:ppa_vs_glo}. In our work, we adhere to the linear evaluation protocol and use a simple logistic regression classifier in evaluating word representations \cite{arora2019simple, gupta2020psif}, whereas \citep{mu2018allbutthetop} use a neural network as their classifier. The \colorbox{red!50}{\textbf{red}} colored cells in Table \ref{tab:ppa_vs_glo} denote the cases where the original embeddings outperformed their Post Processed (PPA) counterparts. Such cases occurred in $14$ out of $27$ comparisons in Table \ref{tab:ppa_vs_glo}. The results in Table \ref{tab:ppa_vs_glo} show that post-processing doesn't always lead to accuracy gains and can be counterproductive in a number of tasks. 

\vspace{0.5em}
\noindent \textbf{Analysis: }The results in Table \ref{tab:ppa_vs_glo} are contrary to the expectation that pushing the word embeddings towards isotropy would lead to better downstream performance. This suggests that within the context of downstream sentence classification tasks, projecting word vectors away from the top components leads to a loss of `useful' information. To explain this loss of `useful' information, we could use the analysis from Figure \ref{fig:full_an}. From Figure \ref{fig:full_an}, it is evident that the top dimensions also contain syntactic information, the loss of which adversely impacts downstream classification tasks, which by construction, benefit from both semantic and syntactic information. Also, by just removing the mean (no top component nullification as in PPA), we notice almost zero change in performance for most of the sentence classification tasks in Table \ref{tab:ppa_vs_glo} (the highest change was for TREC, of $-0.4$, still quite low when compared to $-4.4$ for PPA), which demonstrably shows that removing the mean must be ruled out as the possible cause for the drop in classification accuracies.

\vspace{0.5em}
On the same tasks, we also observe a drop in sentence classification accuracy ($2.37$, $1.99$, $3.94$ average drop on word2vec, Glove, fastText respectively) using $150$ dimensional embeddings obtained from PPA based dimensionality reduction \cite{raunak}. This shows that the variance based post-processing algorithms such as PPA \cite{mu2018allbutthetop} and PPA-PCA \cite{raunak}, when used in \emph{downstream tasks} have significant limitations, which could be attributed to the loss of syntactic information.

\subsection{Machine Translation}
\label{sec:5.2}
Recently, \cite{qi2018and} have shown that pre-trained embeddings lead to significant gains in performance for the translation of three low resource languages namely, Azerbaijani (AZ), Belarusian (BE) and Galician (GL) into English (EN). Here, we demonstrate the impact of the post processing algorithm on machine translation (MT) tasks. We replicate the experimental settings of \cite{qi2018and} and use a standard $1$ layer encoder-decoder model with attention \cite{bahdanau2014neural} and a beam size of $5$. Prior to training, we initialize the encoder with fastText word embeddings (no other embeddings are publically available for these languages) trained on Wikipedia\footnote{\url{https://bit.ly/2WkHQ0Y}}. We then use PPA on the pre-trained embeddings and train again. The results of the experiments are presented in Table \ref{tab:pef_mt}. 

\vspace{0.5em}
\noindent \textbf{Analysis: }From the results, it is evident that removing the top principal component(s) leads to a consistent drop in BLEU scores across the three language pairs. The observations are consistent with the previous section, in that removing top components hurts performance in non-similarity based tasks. This can again be explained using the analysis from earlier section i.e. instead of strengthening the embeddings, removing the top components leads to a loss of `useful' information for the Machine translation task. Further, similar to the previous section, we can specifically attribute the performance drop to the loss of syntactic information, since the top components are at least as equally important for syntactic information as the other components, thus, nullifying them hurts performance.

\begin{table}[!htbp]
\centering
\caption{BLEU scores over three different low-resource language pairs with pretrained emebddings and Top D components removed using PPA. \colorbox{green!30}{\textbf{Green}} cells denotes top scores.}
\scalebox{0.9}{
\begin{tabular}{c|c|c|c}
\hline
& \textbf{AZ-\textgreater{}EN} & \textbf{BE-\textgreater{}EN} & \textbf{GL-\textgreater{}EN} \\ \hline
\textbf{Pre-Trained} & \cellcolor{green!30}\textbf{3.24}       & \cellcolor{green!30}\textbf{6.09}       & \cellcolor{green!30}\textbf{15.91}      \\ \hline
\textbf{PPA (D = 1)}   & 3.19                & 6.02                & 14.81               \\ 
\textbf{PPA (D = 2)}   & 3.07                & 5.50                & 13.88               \\ 
\textbf{PPA (D = 3)}  & 3.04                & 5.26                & 13.27               \\ 
\textbf{PPA (D = 4)}   & 2.92                & 4.75                & 13.24               \\ \hline
\end{tabular}}
\label{tab:pef_mt}
\vspace{-1.0em}
\end{table}

\subsection{Summary and Discussion}
To summarize our experiments on variance based post-processing, we conclude the following:
\begin{enumerate}
\vspace{-0.20em}
\item We can not rely on principal components for manipulating word embeddings as freely as the current literature suggests. While eliminating the `common parts' helps improve the discriminativeness between the word embeddings (thereby refining the word similarity scores), pushing the embeddings towards angular isotropy does not lead to performance gain in downstream tasks, e.g. sentence classification and machine translation. Although, we did not assume any generative model for the embeddings in any of the explanations (unlike \cite{arora2016latent}, which makes use of the isotropy assumption to explain empirical observations in factorizing the PMI matrix), our work further casts doubt on the isotropy assumption for word embeddings and suggests that non-isotropy may be integral to performance on downstream tasks.
\vspace{-0.20em}
\item Furthermore, worse performance in non-similarity tasks can be attributed to the loss of syntactic information contained in the top components, suggesting that the specific geometry created through the `common parts' is integral to embeddings capturing syntactic properties. Establishing a link between the syntactic properties of the embedding space and its non-isotropy would be an interesting direction to explore for future work.
\end{enumerate}

\section{Related Work}
Due to the widespread utility of word embeddings, a number of recent works have explored further improving the embeddings post-hoc, as well as trying to better understand and manipulate the geometry of the embedding space. 

\vspace{0.5em}
\noindent \textbf{Post-Processing Word Embeddings} A number of recent works have been proposed to enhance word embedding quality post-hoc \citep{mrkvsic2016counter, faruqui2014retrofitting, mu2018allbutthetop}. Their applications range from better modeling semantic similarities, improving downstream classification performance to dimensionality reduction of the embeddings \citep{raunak}.

\vspace{0.5em}
\noindent \textbf{Word Embedding Geometry} The linear algebraic structure emergent in word embeddings has received considerable attention \citep{allen2019analogies, arora2018linear}, and theoretical links have been established between neural embedding algorithms and factorization based techniques \citep{levy2014neural}. Another prominent line of work has been along the direction of probing tasks \citep{conneau2018senteval}, which use proxy classification tasks to comparatively measure the presence of certain syntactic/semantic properties in the embedding space.

\vspace{0.5em}
Our work focuses on the dimensional properties of the embedding space in the principal component basis, and also analyzes a few post-processing algorithms, thus contributing to the existing literature on both the areas of embedding analysis.

\section{Conclusion and Future Work}
\label{sec:6}
To conclude, besides elucidating redundancy in the word embedding space, we demonstrate that the variance explained by the word embeddings' principal components is not a reliable proxy for the downstream utility of the corresponding representations and that the syntactic information captured by a principal component does not depend on the amount of variance it explains. Further, we show that variance based post-processing algorithms such as PPA is not suitable for tasks which rely more on syntax, such as sentence classification and machine translation. Going further, we wish to explore whether the geometric intuitions developed in our work could be leveraged for contextualized embeddings such as ElMo \cite{peters2018deep}, BERT \cite{devlin2018bert}, and Roberta \cite{liu2019roberta}, etc.

\bibliography{main}

\begin{thebibliography}{27}
\expandafter\ifx\csname natexlab\endcsname\relax\def\natexlab#1{#1}\fi

\bibitem[{Agirre et~al.(2012)Agirre, Diab, Cer, and
  Gonzalez-Agirre}]{agirre2012semeval}
Eneko Agirre, Mona Diab, Daniel Cer, and Aitor Gonzalez-Agirre. 2012.
\newblock Semeval-2012 task 6: A pilot on semantic textual similarity.
\newblock In \emph{Proceedings of the First Joint Conference on Lexical and
  Computational Semantics-Volume 1: Proceedings of the main conference and the
  shared task, and Volume 2: Proceedings of the Sixth International Workshop on
  Semantic Evaluation}, pages 385--393. Association for Computational
  Linguistics.

\bibitem[{Allen and Hospedales(2019)}]{allen2019analogies}
Carl Allen and Timothy Hospedales. 2019.
\newblock Analogies explained: Towards understanding word embeddings.
\newblock In \emph{International Conference on Machine Learning}, pages
  223--231.

\bibitem[{Arora et~al.(2016)Arora, Li, Liang, Ma, and
  Risteski}]{arora2016latent}
Sanjeev Arora, Yuanzhi Li, Yingyu Liang, Tengyu Ma, and Andrej Risteski. 2016.
\newblock A latent variable model approach to pmi-based word embeddings.
\newblock \emph{Transactions of the Association for Computational Linguistics},
  4:385--399.

\bibitem[{Arora et~al.(2018)Arora, Li, Liang, Ma, and
  Risteski}]{arora2018linear}
Sanjeev Arora, Yuanzhi Li, Yingyu Liang, Tengyu Ma, and Andrej Risteski. 2018.
\newblock Linear algebraic structure of word senses, with applications to
  polysemy.
\newblock \emph{Transactions of the Association for Computational Linguistics},
  6:483--495.

\bibitem[{Arora et~al.(2017)Arora, Liang, and Ma}]{arora2016simple}
Sanjeev Arora, Yingyu Liang, and Tengyu Ma. 2017.
\newblock A simple but tough-to-beat baseline for sentence embeddings.
\newblock \emph{International Conference of Learning Representation}.

\bibitem[{Arora et~al.(2019)Arora, Liang, and Ma}]{arora2019simple}
Sanjeev Arora, Yingyu Liang, and Tengyu Ma. 2019.
\newblock A simple but tough-to-beat baseline for sentence embeddings.
\newblock In \emph{5th International Conference on Learning Representations,
  ICLR 2017}.

\bibitem[{Bahdanau et~al.(2015)Bahdanau, Cho, and Bengio}]{bahdanau2014neural}
Dzmitry Bahdanau, Kyunghyun Cho, and Yoshua Bengio. 2015.
\newblock Neural machine translation by jointly learning to align and
  translate.
\newblock In \emph{3rd International Conference on Learning Representations,
  ICLR 2015}.

\bibitem[{Bojanowski et~al.(2017)Bojanowski, Grave, Joulin, and
  Mikolov}]{bojanowski2017enriching}
Piotr Bojanowski, Edouard Grave, Armand Joulin, and Tomas Mikolov. 2017.
\newblock Enriching word vectors with subword information.
\newblock \emph{Transactions of the Association of Computational Linguistics},
  5(1):135--146.

\bibitem[{Conneau and Kiela(2018)}]{conneau2018senteval}
Alexis Conneau and Douwe Kiela. 2018.
\newblock Senteval: An evaluation toolkit for universal sentence
  representations.
\newblock In \emph{Proceedings of the Eleventh International Conference on
  Language Resources and Evaluation (LREC 2018)}.

\bibitem[{Conneau et~al.(2018)Conneau, Kruszewski, Lample, Barrault, and
  Baroni}]{conneau2018you}
Alexis Conneau, German Kruszewski, Guillaume Lample, Loic Barrault, and Marco
  Baroni. 2018.
\newblock What you can cram into a single vector: Probing sentence embeddings
  for linguistic properties.
\newblock In \emph{Proceedings of the 56th Annual Meeting of the Association
  for Computational Linguistics (Volume 1: Long Papers)}, pages 2126--2136.

\bibitem[{Devlin et~al.(2019)Devlin, Chang, Lee, and
  Toutanova}]{devlin2018bert}
Jacob Devlin, Ming-Wei Chang, Kenton Lee, and Kristina Toutanova. 2019.
\newblock {BERT}: Pre-training of deep bidirectional transformers for language
  understanding.
\newblock In \emph{Proceedings of the 2019 Conference of the North {A}merican
  Chapter of the Association for Computational Linguistics: Human Language
  Technologies, Volume 1 (Long and Short Papers)}, pages 4171--4186,
  Minneapolis, Minnesota. Association for Computational Linguistics.

\bibitem[{Faruqui et~al.(2014)Faruqui, Dodge, Jauhar, Dyer, Hovy, and
  Smith}]{faruqui2014retrofitting}
Manaal Faruqui, Jesse Dodge, Sujay~K Jauhar, Chris Dyer, Eduard Hovy, and
  Noah~A Smith. 2014.
\newblock Retrofitting word vectors to semantic lexicons.
\newblock \emph{arXiv preprint arXiv:1411.4166}.

\bibitem[{Faruqui et~al.(2015)Faruqui, Dodge, Jauhar, Dyer, Hovy, and
  Smith}]{faruqui2015retrofitting}
Manaal Faruqui, Jesse Dodge, Sujay~Kumar Jauhar, Chris Dyer, Eduard Hovy, and
  Noah~A Smith. 2015.
\newblock Retrofitting word vectors to semantic lexicons.
\newblock In \emph{Proceedings of the 2015 Conference of the North American
  Chapter of the Association for Computational Linguistics: Human Language
  Technologies}, pages 1606--1615.

\bibitem[{Faruqui and Dyer(2014)}]{faruqui2014community}
Manaal Faruqui and Chris Dyer. 2014.
\newblock Community evaluation and exchange of word vectors at wordvectors.
  org.
\newblock In \emph{Proceedings of 52nd Annual Meeting of the Association for
  Computational Linguistics: System Demonstrations}, pages 19--24.

\bibitem[{Gupta et~al.(2020)Gupta, Saw, Nokhiz, Netrapalli, Rai, and
  Talukdar}]{gupta2020psif}
Vivek Gupta, Ankit Saw, Pegah Nokhiz, Praneeth Netrapalli, Piyush Rai, and
  Partha Talukdar. 2020.
\newblock P-sif: Document embeddings using partition averaging.
\newblock In \emph{Proceedings of the AAAI Conference on Artificial
  Intelligence}.

\bibitem[{Jolliffe and Cadima(2016)}]{jolliffe2016principal}
Ian~T Jolliffe and Jorge Cadima. 2016.
\newblock Principal component analysis: a review and recent developments.
\newblock \emph{Philosophical Transactions of the Royal Society A:
  Mathematical, Physical and Engineering Sciences}, 374(2065):20150202.

\bibitem[{Levy and Goldberg(2014)}]{levy2014neural}
Omer Levy and Yoav Goldberg. 2014.
\newblock Neural word embedding as implicit matrix factorization.
\newblock In \emph{Advances in neural information processing systems}, pages
  2177--2185.

\bibitem[{Liu et~al.(2019)Liu, Ott, Goyal, Du, Joshi, Chen, Levy, Lewis,
  Zettlemoyer, and Stoyanov}]{liu2019roberta}
Yinhan Liu, Myle Ott, Naman Goyal, Jingfei Du, Mandar Joshi, Danqi Chen, Omer
  Levy, Mike Lewis, Luke Zettlemoyer, and Veselin Stoyanov. 2019.
\newblock Roberta: A robustly optimized bert pretraining approach.
\newblock \emph{arXiv preprint arXiv:1907.11692}.

\bibitem[{Mikolov et~al.(2013)Mikolov, Sutskever, Chen, Corrado, and
  Dean}]{mikolov2013distributed}
Tomas Mikolov, Ilya Sutskever, Kai Chen, Greg~S Corrado, and Jeff Dean. 2013.
\newblock Distributed representations of words and phrases and their
  compositionality.
\newblock In \emph{Advances in neural information processing systems}, pages
  3111--3119.

\bibitem[{Mimno and Thompson(2017)}]{mimno2017strange}
David Mimno and Laure Thompson. 2017.
\newblock The strange geometry of skip-gram with negative sampling.
\newblock In \emph{Proceedings of the 2017 Conference on Empirical Methods in
  Natural Language Processing}, pages 2873--2878.

\bibitem[{Mrk{\v{s}}ic et~al.()Mrk{\v{s}}ic, OS{\'e}aghdha, Thomson, and
  Ga{\v{s}}i{\'{c}, Milica and Rojas-Barahona, Lina and Su, Pei-Hao and
  Vandyke, David and Wen, Tsung-Hsien and Young, Steve}}]{mrkvsic2016counter}
Nikola Mrk{\v{s}}ic, Diarmuid OS{\'e}aghdha, Blaise Thomson, and
  pages={142--148}~year={2016} Ga{\v{s}}i{\'{c}, Milica and Rojas-Barahona,
  Lina and Su, Pei-Hao and Vandyke, David and Wen, Tsung-Hsien and Young,
  Steve}, booktitle={Proceedings of NAACL-HLT}.
\newblock Counter-fitting word vectors to linguistic constraints.

\bibitem[{Mu and Viswanath(2018)}]{mu2018allbutthetop}
Jiaqi Mu and Pramod Viswanath. 2018.
\newblock All-but-the-top: Simple and effective postprocessing for word
  representations.
\newblock In \emph{International Conference on Learning Representations}.

\bibitem[{Pennington et~al.(2014)Pennington, Socher, and
  Manning}]{pennington2014glove}
Jeffrey Pennington, Richard Socher, and Christopher Manning. 2014.
\newblock Glove: Global vectors for word representation.
\newblock In \emph{Proceedings of the 2014 conference on empirical methods in
  natural language processing (EMNLP)}, pages 1532--1543.

\bibitem[{Peters et~al.(2018)Peters, Neumann, Iyyer, Gardner, Clark, Lee, and
  Zettlemoyer}]{peters2018deep}
Matthew Peters, Mark Neumann, Mohit Iyyer, Matt Gardner, Christopher Clark,
  Kenton Lee, and Luke Zettlemoyer. 2018.
\newblock Deep contextualized word representations.
\newblock In \emph{Proceedings of the 2018 Conference of the North American
  Chapter of the Association for Computational Linguistics: Human Language
  Technologies, Volume 1 (Long Papers)}, pages 2227--2237.

\bibitem[{Qi et~al.(2018)Qi, Sachan, Felix, Padmanabhan, and
  Neubig}]{qi2018and}
Ye~Qi, Devendra Sachan, Matthieu Felix, Sarguna Padmanabhan, and Graham Neubig.
  2018.
\newblock When and why are pre-trained word embeddings useful for neural
  machine translation?
\newblock In \emph{Proceedings of the 2018 Conference of the North American
  Chapter of the Association for Computational Linguistics: Human Language
  Technologies, Volume 2 (Short Papers)}, pages 529--535.

\bibitem[{Raunak et~al.(2019)Raunak, Gupta, and Metze}]{raunak}
Vikas Raunak, Vivek Gupta, and Florian Metze. 2019.
\newblock Effective dimensionality reduction for word embeddings.
\newblock In \emph{Proceedings of the 4th Workshop on Representation Learning
  for NLP (RepL4NLP-2019)}, pages 235--243, Florence, Italy. Association for
  Computational Linguistics.

\bibitem[{Yin and Shen(2018)}]{dim}
Zi~Yin and Yuanyuan Shen. 2018.
\newblock On the dimensionality of word embedding.
\newblock In S.~Bengio, H.~Wallach, H.~Larochelle, K.~Grauman, N.~Cesa-Bianchi,
  and R.~Garnett, editors, \emph{Advances in Neural Information Processing
  Systems 31}, pages 895--906. Curran Associates, Inc.

\end{thebibliography}
\bibliographystyle{acl_natbib}

\end{document}